\documentclass[conference]{IEEEtran}
\IEEEoverridecommandlockouts
\usepackage[english]{babel}
\usepackage{booktabs}
\usepackage{multirow}
\usepackage{graphicx}
\usepackage{inconsolata}
\usepackage{dirtytalk}
\usepackage{amsmath}
\usepackage{bbding}
\usepackage{amssymb}
\usepackage{pifont}

\usepackage{caption} 
\usepackage{cite}
\usepackage{amsmath,amssymb,amsfonts}
\usepackage{algorithmic}
\usepackage{graphicx}
\usepackage{textcomp}
\usepackage{xcolor}

\usepackage{verbatim}

\usepackage{hyperref}

\def\BibTeX{{\rm B\kern-.05em{\sc i\kern-.025em b}\kern-.08em
    T\kern-.1667em\lower.7ex\hbox{E}\kern-.125emX}}
\DeclareUnicodeCharacter{202F}{\,}

\begin{document}

\title{ End-to-End  $n$-ary Relation Extraction \\ for Combination Drug Therapies\\

\thanks{Research reported in this paper was supported by the National Library of Medicine of the National Institutes of Health (NIH) under Award Number R01LM013240. The content is solely the responsibility of the authors and does not necessarily represent the official views of the NIH.}
}

\author{\IEEEauthorblockN{ Yuhang Jiang}
\IEEEauthorblockA{\textit{Department of Computer Science} \\
\textit{University of Kentucky}\\
Lexington, KY USA\\
yuhang.jiang@uky.edu}
\and
\IEEEauthorblockN{Ramakanth Kavuluru}
\IEEEauthorblockA{\textit{Division of Biomedical Informatics} \\
\textit{Dept.~of Internal Medicine, Univ.~of Kentucky}\\
Lexington, KY USA \\
ramakanth.kavuluru@uky.edu}
}

\maketitle

\begin{abstract}
Combination drug therapies are treatment regimens that involve two or more drugs, administered more commonly for patients with cancer, HIV, malaria, or tuberculosis. Currently there are over 350K articles in PubMed that use the {\color{blue} \texttt{combination drug therapy}} MeSH heading with at least 10K articles published per year over the past two decades. Extracting combination therapies from scientific literature inherently constitutes an  $n$-ary relation extraction problem. Unlike in the general $n$-ary setting where $n$ is fixed (e.g., drug-gene-mutation relations where $n=3$),  extracting combination therapies is a special setting where $n \geq 2$ is dynamic, depending on each instance. Recently, Tiktinsky~et al. (NAACL 2022) introduced a first of its kind dataset, \texttt{CombDrugExt}, for extracting such therapies from literature. Here, we use a sequence-to-sequence style end-to-end extraction method to achieve an F1-Score of $66.7\%$ on the \texttt{CombDrugExt} test set for positive (or effective) combinations. This is an absolute $\approx 5\%$ F1-score improvement even over the prior best relation classification  score with spotted drug entities (hence, not end-to-end). Thus our effort introduces a state-of-the-art first model for end-to-end extraction that is already superior to the best prior non end-to-end model for this task. Our model seamlessly extracts all drug entities and relations in a single pass and is highly suitable for dynamic $n$-ary extraction scenarios. 

\end{abstract}


\section{Introduction}
Combination drug therapies (CDTs) have been on the rise for serious conditions such as cancers and HIV. Nearly half of the papers on PubMed that are tagged with the \texttt{combination drug therapy} MeSH heading are related to cancer treatment. There are at least 10K articles published every year discussing advances in combination drug therapies. As such, there is a need for automatically extracting and creating structured knowledge bases of such therapies. A CDT involves at least two drugs that are used as part of a treatment regimen for a specific disease.\footnote{We distinguish this from  \texttt{drug combination} (a different MeSH term)  typically employed to denote pills or formulations that have two are more ingredients in the same product (e.g., Vicks NyQuil has both acetaminophen and dextromethorphan.) We, however, use the term ``combination'' in this paper to mean a drug set that is part of a combination therapy regimen.}

As CDTs become more prominent due to advances in precision medicine strategies~\cite{klauschen2014combinatorial}, it is important to extract new combinations as reported in scientific literature. However, until recently, we are not aware of any public datasets   to extract CDTs from textual data, with the eventual goal of curating large sets of such therapies from scientific literature. The first such dataset and associated extraction baselines were presented by Tiktinsky et al.~\cite{tiktinsky-etal-2022-dataset} in 2022. They manually annotated 1600 abstracts with different drug combinations and provided train-validation-test splits for bench marking by other researchers. For the rest of this manuscript we call this dataset \textbf{\texttt{CombDrugExt}}. 
We describe the dataset characteristics and the specifics of the end-to-end $n$-ary extraction task in Section~\ref{sec-data}. 

To begin with, it is straightforward to see that CDT extraction  from text is inherently an $n$-ary relation extraction task, where $n \geq 2$ is dynamic and changes based on each input instance. This is unlike other $n$-ary relation extraction tasks where $n$ is typically fixed (e.g., drug-gene-mutation relation extraction task where $n=3$). 
A simple example of an effective two-drug combination therapy can be seen from the \textbf{bold} tokens in the sentence --- ``\textbf{Abiraterone acetate} in combination with \textbf{prednisone} was the first approved hormone therapy demonstrating survival benefit in metastatic castration-resistant prostate cancer.'' Likewise, a simple negative example of a lack of combination for the two drugs mentioned can be noticed from the {\color{red}red} tokens in this sentence  --- ``{\color{red}Atorvastatin} and {\color{red}simvastatin} (members of the 3-hydroxy-3-methylglutaryl coenzyme A reductase inhibitor family) are widely prescribed as cholesterol-lowering agents.'' At a high level, the extraction challenge is to spot the drugs and then distinguish drug sets that are discussed as a combination therapy from drug sets that are discussed in other ways (e.g., part of a list or a comparative statement). As the number of drugs in an instance increases, the difficulty also increases in terms of identifying groups of drugs that form a combination compared with other groups that are simply discussed in   ways that do not convey a combination.

Since $n$ is not known upfront for the task (it ranges from 2 to 15 in the \textbf{\texttt{CombDrugExt}} dataset), we need an extraction approach that dynamically adjusts the grouping without resorting to brute force enumeration, which could quickly lead to an expensive combinatorial explosion of the candidate combination drug set space. This rules out the typical pipeline approach of recognizing all drugs through named entity recognition (NER) first and then checking if subsets from the list identified represent a CDT. 
Toward more efficient single-pass CDT extraction,  we adapt the sequence-to-sequence style relation extraction approach recently introduced by Giorgi et al.~\cite{giorgi-etal-2022-sequence}, called Seq2Rel. 

Sequence-to-sequence models have been originally popularized for machine translation. This is typically carried out in an encoder-decoder  architecture where the encoder processes the source sentence with the encoder output used by the decoder to output tokens, in order, of the target sentence. This encoder-decoder setup has been shown to be powerful and flexible enough to represent relation extraction tasks. This was found to be specifically more advantageous in end-to-end settings where negative examples are not explicitly annotated in training data and in $n$-ary extraction scenarios. The general idea is to have the input sentence processed by the encoder, whose output is then used by the decoder to generate relations as a target sequence.
This is done through (1)~a so called \textit{linearization schema} that helps  represent and interpret the output sequence as relations (more in Section~\ref{sec-methods}) and (2)~a copy-mechanism where the decoder is constrained to output only tokens that are observed in the sequence input to the encoder (unlike from the full vocabulary of the target language in  machine translation). This enables the decoder to output spans of the input text that correspond to entities and special tokens that correspond to relation labels connecting entities. 

With this introductory setup, our main contributions in CDT extraction can be summarized as follows 
\begin{itemize}
\item We adapt the Seq2Rel~\cite{giorgi-etal-2022-sequence} approach through a  linearization schema that fits the end-to-end CDT extraction task  for the \textbf{\texttt{CombDrugExt}} dataset~\cite{tiktinsky-etal-2022-dataset}. We show that a model built with this architecture achieves an end-to-end F1-score that is $\approx  5\%$ better than the \textbf{non} end-to-end prior best score for effective drug combination extraction. 
\item We also develop a modified linearization schema for Seq2Rel that not only outputs relations but also all entities in an exhaustive manner (regardless of whether they participate in a relation). Though this was not a proposed task for \textbf{\texttt{CombDrugExt}}, we show that this approach not only helps with drug NER, but also leads to better  overall end-to-end relation extraction performance  when considering both effective and non-effective combinations. 
\item We examine patterns in   false positives and false negatives through error analysis. We also study how context length affects performance in this approach through simple ablation experiments. 
\end{itemize}
The code to  reproduce our results is available here: \url{https://github.com/bionlproc/end-to-end-CombDrugExt}. The original dataset is already publicly available: \url{https://github.com/allenai/drug-combo-extraction}.

\begin{table*}[t]
\renewcommand{\arraystretch}{1.2}
\centering
\caption{Examples of different drug combination relations indicating sentences and corresponding linearized model outputs.}

\label{fig:complexities}

\begin{tabular}{p{0.15\textwidth} p{0.55\textwidth} p{0.22\textwidth}}
\toprule
Relation Classes &
  Example &
  Comment \\ \midrule
  \multirow{3.5}{*}{Positive Combination  } &
  INPUT: Codelivery of \textbf{sorafenib} and \textbf{curcumin} by directed self-assembled nanoparticles enhances therapeutic effect on hepatocellular carcinoma. &
  \multirow{3.5}{*}{\parbox{4cm}{{\textbf{sorafenib} and \textbf{curcumin}} are an effective combination.}} \\ \cmidrule(lr){2-2}
 &
  OUTPUT: \texttt{{\textbf{sorafenib} @DRUG@} {\textbf{curcumin} @DRUG@ @POS@}}  &
   \\ \cmidrule(r){1-3}
  \multirow{3.5}{*}{Non-pos. Combination} &
  INPUT: Patients received \textbf{docetaxel} 35 mg/m(2) and \textbf{irinotecan} 60 mg/m(2), intravenously, on Days 1 and 8, every 21 days, until disease progression. &
  \multirow{3.5}{*}{\parbox{4cm}{A combination (\textbf{docetaxel} and \textbf{irinotecan}) is employed with no evidence of effectiveness.}} \\ \cmidrule(lr){2-2}
 &
   OUTPUT: \texttt{{ \textbf{docetaxel}}  { @DRUG@} { \textbf{irinotecan} @DRUG@} @COMB@} &
   \\ \cmidrule(r){1-3}
   
\multirow{3.5}{*}{Not a Combination} &
  INPUT: The results showed that \textbf{lamotrigine} did not produce any change in cognitive function, while \textbf{carbamazepine} produced cognitive dysfunction. &
  \multirow{3.5}{*}{\parbox{4cm}{\textbf{lamotrigine} and \textbf{carbamazepine} are used separately.}} \\ \cmidrule(lr){2-2}
 &
   OUTPUT: \texttt{{ \textbf{lamotrigine} @DRUG@} { \textbf{carbamazepine} @DRUG@} @NOCOMB@} &
   \\ \bottomrule
\end{tabular}
\vspace{-3mm}
\end{table*}

\section{The \textbf{\texttt{CombDrugExt}}  Dataset}
\label{sec-data}
 In this section we outline the characteristics of the \textbf{\texttt{CombDrugExt}} dataset, the associated  $n$-ary combination extraction task as designed by Tiktinsky et al.~\cite{tiktinsky-etal-2022-dataset}, and the end-to-end variation we consider in this paper.
 
The full dataset consists of 1600 manually annotated abstracts, each of which mentions between 2 and 15 drugs. 840 of these abstracts describe one or more drug combinations that have a positive effect, with the number of drugs in each such combination varying from 2 to 11. The remaining 760 abstracts either contain mentions of drugs that are not used in combination or discuss combinations of drugs that do not have a combined positive effect. The abstracts (with associated annotations) are subdivided into training, validation, and test splits, which we use for our experiments and evaluation.

\subsection{CDT Relation classes}
\label{sec-classes}
The relations are categorized into three classes: (a).~``Positive combination" (\texttt{POS}), where the text suggests that the drugs are used together in a treatment and are described or implied to have a positive effect; (b).~``Non-positive combination" (\texttt{COMB}), where the input sentence indicates that the drugs are used together in a treatment, but there is no evidence in the text to suggest that it is effective; (c).~``Not a combination" (\texttt{NOCOMB}), referring to an instance where there is no indication in the text to suggest that the drugs are used together at all.
An example for each class is shown in Table~\ref{fig:complexities} (along with the linearized output representations for training/inference, which we discuss a little later in the manuscript). 
At this juncture it is important to state two important aspects of the task:
\begin{enumerate}
\item The annotation (and extraction) task setup by  Tiktinsky et al.~\cite{tiktinsky-etal-2022-dataset}  does not involve identifying the target condition or disease for which the combination is being considered. 
This target is only relevant for the \texttt{POS} and \texttt{COMB} relations as there isn't a notion of a target for a set of drugs that do not even participate in a combination (i.e., for the \texttt{NOCOMB} class). 
Even though this is suboptimal, we stick with this task definition introduced by Tiktinsky et al.~\cite{tiktinsky-etal-2022-dataset} in the rest of this paper. We intend to extend \textbf{\texttt{CombDrugExt}} and the associated task to also identify the target disease for \texttt{POS} and \texttt{COMB} relations in a future effort. For example, the first sentence in Table~\ref{fig:complexities} would have the target attribute of ``hepatocellular carcinoma.''
\item Each input instance is provided as an abstract within which a \textbf{single sentence} that has all the drugs involved is designated for extraction. As such, this is a sentence level end-to-end extraction task, with the provision of a broader abstract context which contains the sentence\footnote{We recognize that this is not   the most generic formulation; ideally, all entities should not be required to be present in the same sentence. However, considering the setup in the \textbf{\texttt{CombDrugExt}} dataset, we make this assumption.}. Models ought to process the sentence and may optionally consider the context surrounding it in the full abstract. Multiple combinations are allowed as output for each sentence.
We note that Tiktinsky et al.~\cite{tiktinsky-etal-2022-dataset} assume that the entities are already spotted and hence solve the relation classification problem; we propose an end-to-end solution that operates on input text without known drug spans.
\end{enumerate}

\subsection{Formal end-to-end task definition}
\label{sec-formal}
Succinctly, given a sentence that contains drug names and an enclosing context (here, abstract), the end-to-end CDT extraction task is to identify all stated sets of drug  spans that correspond to either a \texttt{POS} or \texttt{COMB} relation; for sentences where no such relations exist, output the set of all drug mentions with a \texttt{NOCOMB} label.   

More formally, the task is to consider an input instance $X=(C, i)$  where $C=[S_1,\ldots,S_n]$ is an ordered list of $n$ sentences (here, all the sentences in an abstract) and $1 \leq i \leq n$ is an index of a  sentence $S_i$ that is designated as the input sentence.
Let  $D (S_i) =\{(d^1_{start}, d^1_{end}), \ldots, (d^m_{start}, d^m_{end})\}$ be the set of $m>=2$ spans of drug mentions in $S_i$ (which is not part of the input instance in our end-to-end assumption). The extraction model  output should  be a result  set $R(X) =\{(A_j, y_j)\}$ where $A_j \in \mathcal{P}(D(S_i))$ is a drug combination from the power set of $D(S_i)$  and $y_j \in \{ \texttt{POS}, \texttt{COMB}\}$. When $X$ does not contain any combinations, the model output is simply the singleton $R(X) = \{ (D(S_i), \texttt{NOCOMB})\}$.

\subsection{Data and task characteristics} \label{challenges}
There are already few $n$-ary tasks in general in biomedical relation extraction as the focus tends to be more on binary relations, which is reasonable. As indicated in prior sections, the dynamic nature of $n$ in this task is unique and it becomes trickier as drugs that are part of a combination and other drugs that are mentioned in a different sense may all occur in the same input sentence. The model ought to separate these two sets and may also need to identify multiple \texttt{POS}/\texttt{COMB} relations within the same sentence --- 16\% of sentences containing drug combinations in \textbf{\texttt{CombDrugExt}}  have more than one combination.  Around 70\% of \texttt{POS} combinations are binary, 19\% are $3$-ary, and over 5\% are $4$-ary\footnote{Although these were counted to throw light on data characteristics, the proportions were not used in designing or evaluating the model.}. Although the task is in one sense sentence level, the enclosing context of the full abstract is essential in several cases. The drug combination and the evidence whether it is effective may be far apart in the abstract. In fact, in an extreme case, sentences containing the effectiveness evidence and the participating drug names were separated by 41 sentences. Other linguistic phenomena including coordination, numerical reasoning, and (biomedical) world knowledge may also need to be considered in arriving at combinations.

\section{End-to-End Model for CDT Extraction}
\label{sec-methods}
In this effort, we use the sequence-to-sequence neural architecture Seq2Rel by Giorgi et al.~\cite{giorgi-etal-2022-sequence}, which was designed for flexible end-to-end relation extraction. Compared with pipeline architectures that are particularly expensive for the dynamic $n$-ary nature of CDT extraction, Seq2Rel can be adapted easily to extract  relations (and involved entities) in a single pass. Recall that our input sequence is a sentence with an enclosing full abstract context and the output is a list of 2-tuples each of which is a set of set of drugs and the corresponding label (\texttt{POS}, \texttt{COMB}, or \texttt{NOCOMB}). Before we describe the architecture, we first present the representation of these output 2-tuples. 

\begin{figure*}[t!]
    \centering
    \includegraphics[scale=0.45]{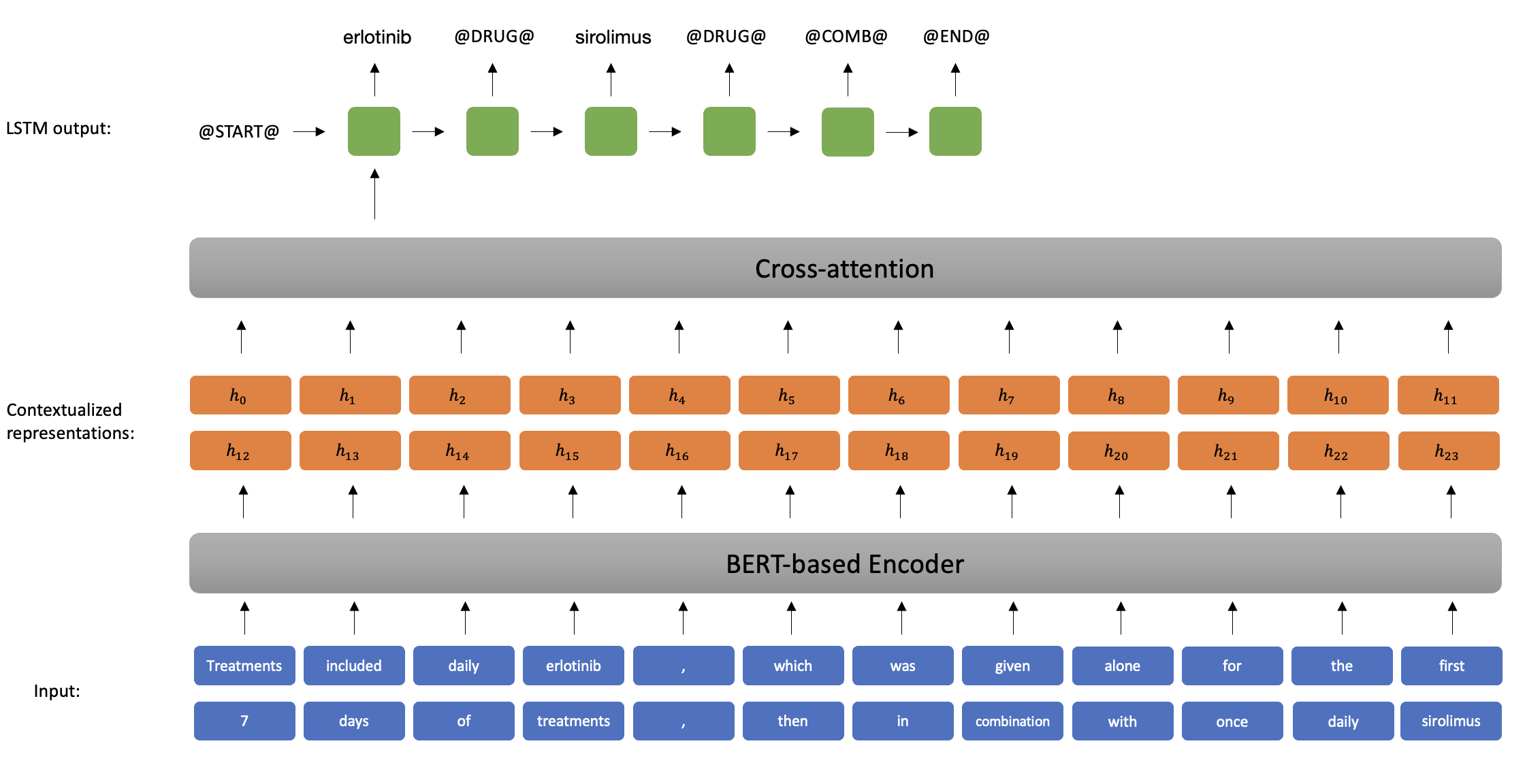}
    \caption{The Seq2Rel architecture for CDT extraction with an input sentence and a serialized output of drug combinations}
    \label{fig:model}
\end{figure*}

\subsection{Linearization schema for CDT} \label{linearization}
A sentence is easy and natural to represent as a sequence. However, representing 2-tuples that correspond to the output for the   end-to-end task (as described in Section~\ref{sec-formal}) needs a clear specification, which we call the linearization schema. It involves special tokens that represent different relation labels in \textbf{\texttt{CombDrugExt}} (as shown in Table~\ref{tb:stoken}) and the special token @DRUG@ to represent a drug entity.  

\begin{table}[h]
\centering
\renewcommand{\arraystretch}{1.2}
\caption{Special tokens for relations in \textbf{\texttt{CombDrugExt}}}
\begin{tabular}{r ccc}
\toprule
Relation labels & Special token \\\midrule
Positive combination  & @POS@ \\
Non-positive combination   & @COMB@\\
 Not a combination & @NOCOMB@ \\
\bottomrule
\end{tabular}

\label{tb:stoken}
\end{table}

The representation is straightforward in that it lists all the drug spans forming a combination, in the order that was originally provided by Tiktinksky et al.~\cite{tiktinsky-etal-2022-dataset}, each followed by the @DRUG@ token with a suitable combination label token (from Table~\ref{tb:stoken})) as the final token. 
For a simple example shown in the \texttt{POS} relation in Table~\ref{fig:complexities} (Row 1), the linearization is shown right below it where the two drugs spans are followed by the @POS@ token. If more than two drugs are part of the combination, all of them are generated before the label is output. Outputs with more than one combination are also represented the same way with the combination order determined by the ordering provided by dataset creators~\cite{tiktinsky-etal-2022-dataset}.

For instance, consider the sentence: ``In non-metastatic castration-resistant prostate cancer, 
two second-generation anti-androgens, {\textbf{apalutamide} } and {\textbf{enzalutamide}}, when  
used in combination with {\textbf{ADT}}, have demonstrated a significant benefit in 
metastasis-free survival.'' Here the expected output is: {\textbf{apalutamide}} @DRUG@ {\textbf{ADT}} @DRUG@  @POS@ {\textbf{enzalutamide}} @DRUG@ 
{\textbf{ADT}} @DRUG@  @POS@. Here, \textbf{ADT} is used in combination with each of the other two drugs as asserted in the sentence.

\subsection{Model architecture}
The Seq2Rel architecture is shown in  Figure~\ref{fig:model} populated with an example for CDT extraction. A BERT based encoder~\cite{devlin-etal-2019-bert} takes the input sentence (along with accompanying context from the full abstract) and maps each token to a contextual embedding. A single layer LSTM~\cite{HochSchm97} serves as the decoder for generating the serialized output sequence as discussed in Section~\ref{linearization}. The probability $P(Y|X)$ for the serialized representation $Y$ of $R(X)$ (from Section~\ref{sec-formal}) for input $X$ is computed using the chain rule involving probability estimates for prefixes of $Y$ given $X$ as is common in encoder-decoder architectures for sequence-to-sequence modeling. The loss function is the average cross-entropy per target word. We refer the reader to Chapter 9.7 of Jurafsky and Martin~\cite{slpbook3} for further details.

Given this setup is prone to generating arbitrary tokens that do not appear in the input sequence (as the output sequence is typically taken from the entire vocabulary), generation is restricted to special tokens and only tokens from the input sentence through a copy mechanism~\cite{zeng-etal-2018-extracting} as  implemented by Giorgi et al.~\cite{giorgi-etal-2022-sequence}. This allows the decoder to obtain a probability distribution over input and special tokens at each time step, thus ensuring that only expected units are generated in the serialized output.

\subsection{Implementation, training, and post-processing}
\label{sec-post}
We used PubMedBERT~\cite{gu2021domain} as the encoder given it is pretrained from scratch with PubMed abstracts and articles and was also used by the baselines for this task~\cite{giorgi-etal-2022-sequence}. The single layer LSTM with random initialized weights was used. We trained the model for 130 epochs, with a learning rate of 2e-5 for encoder and a learning rate of 1.21e-4 for decoder. We trained the model on Google Colab (\url{https://colab.research.google.com}), which took nearly 2.5 hours.

Due to how WordPiece tokenization works in BERT models, we noticed our model outputs extra spaces around the hyphen (``-'') character in drugs containing it. As an example, consider the sentence:  ``Nal-IRI with \textbf{5-fluorouracil} (5-FU) and leucovorin or gemcitabine plus cisplatin
in advanced biliary tract cancer-the NIFE trial."
Note the extra spaces around hyphen in the serialized output: 
``\textbf{5 - fluorouracil} @DRUG@ leucovorin @DRUG@ @COMB@ gemcitabine @DRUG@ 
cisplatin @DRUG@ @COMB@.''
So we simply post-processed such outputs to remove these extra spaces (around hyphens) to match strings in the input sentence.

\section{Main Experiments and Results}
\label{sec-main-res}
We evaluate our model based on the F1-score (for comparison with prior efforts) while examining both precision and recall. We use the ``exact match" criterion where a relation is correct if (a).~all drugs in the combination are identified and match the gold spans and (b).~the relation label is correct. With this notion of correctness, F1-score is the harmonic mean of 
\begin{equation*}
    Recall = \#correct\_relations/\#all\_gold\_relations \quad \mbox{and}
\end{equation*}
\begin{equation*}
    Precision = \#correct\_relations/\#predicted\_relations.
\end{equation*}

There are two types of evaluations reported in the original effort by Tiktinsky et al.~\cite{tiktinsky-etal-2022-dataset} and we do the same here. The first one is the F1-score for \texttt{POS} class and the second one is a relaxed setup where \texttt{POS} and \texttt{COMB} classes are combined into the same class \texttt{ANY-COMB} and report its F1-score. The natural way to model the original 3-class setup is to use all three tags (from Table~\ref{tb:stoken}), which inherently gives per-class F1-scores. But to compute the \texttt{ANY-COMB} scores, we just collapse the \texttt{POS} and \texttt{COMB} labels to the same \texttt{ANY-COMB} label and compute the \texttt{ANY-COMB} F1-score. 

A slightly different way is to use a 2-class set up during the training and testing through a different labeling scheme upfront. For the  \texttt{POS} classifier, we simply use two tags: @POS@ and @NON-POS@, where the latter is the tag given to instances in the union of \texttt{COMB} and \texttt{NOCOMB}. For the \texttt{ANY-COMB} classifier, we use a new tag @ANY-COMB@ for instances in the union of \texttt{POS} and \texttt{COMB} classes and the @NOCOMB@ is the tag for the other class. 
We show the scores obtained through both the 3-class and 2-class modeling in Table~\ref{2-way}\footnote{Note that the first row of this table is computed using a single 3-way classifier while the second row is based on two binary classifiers.}. We note that both approaches achieved similar scores. The the 3-way setup results in a slightly better \texttt{POS} score and the 2-way labeling improves by $1.6\%$ in \texttt{ANY-COMB} F1-score compared with the 3-class approach.  Precision and recall are also close to each other but overall recall is slightly less than precision.

\begin{table}[h]
\centering
\renewcommand{\arraystretch}{1.4}
\caption{Our main results (precision, recall, F1-score) using 3-way and 2-way classifier modeling for CDT extraction.}
\label{2-way}
\begin{tabular}{lcccccc}
\toprule
\multicolumn{1}{c}{Methods} & \multicolumn{3}{c}{\texttt{POS} Combination} & \multicolumn{3}{c}{Any Combination} \\
 & $F_1$ & P & R & $F_1$ & P & R\\\midrule
3-way classification  & \textbf{66.7} & 68.1 & 65.3 & 71.1 & 73.5 & 68.9 \\
2-way classification & 66.4 & 69.1 & 64.0 & \textbf{72.7} & 74.1 & 71.3  \\
\bottomrule
\end{tabular}

\end{table}

\begin{table*}[!ht] 
  \centering
  \renewcommand{\arraystretch}{1.4}
    \caption{Our results (last row) compared with different baseline foundation models (DAPT means with continued domain-adaptive pretraining~\cite{gururangan-etal-2020-dont}) trained with the PURE method \cite{zhong-chen-2021-frustratingly}. The baseline results (first 5 rows) are from Tiktinsky et al.~\cite{tiktinsky-etal-2022-dataset}.
  }
  \begin{tabular}{lcc} 
  \toprule
  \textbf{Model} & \multicolumn{1}{c}{\textbf{Positive Combination $F_1$}} & \multicolumn{1}{c}{\textbf{Any Combination $F_1$}}\\\midrule
  PURE (w/SciBERT)  & 44.6  & 50.2\\ 
  PURE (w/BlueBERT)  & 41.2  & 47.3\\ 
  PURE (w/BioBERT)  & 45.4  & 46.7\\ 
  PURE (w/PubMedBERT)  & 50.7  & 55.9\\ \bottomrule
  PURE (w/PubMedBERT+DAPT)  & 61.8  & 69.4\\ \bottomrule
  Seq2Rel (w/PubMedBERT) (\textbf{\color{blue}end-to-end})& \textbf{66.7}  & \textbf{71.1}\\
  \bottomrule
  \end{tabular}
   \label{table:main-results}
\end{table*}

Next, we compare our results with prior state-of-the-art  in Table~\ref{table:main-results}, where the last row contains the F1-scores of our 3-way model (from Table~\ref{2-way}).
We  see that for both \texttt{POS} and \texttt{ANY-COMB} settings, our Seq2Rel model has better performance. Compared with the penultimate row, which shows the prior SoTA score on this dataset, we have an almost $5\%$ improvement in \texttt{POS} score even though our model is end-to-end while all other rows in the table assume drugs names are spotted.  All other models are trained with the PURE method~\cite{zhong-chen-2021-frustratingly}, a span-based approach where entity markers are   used to glean signal for relation classification. The gains for \texttt{ANY-COMB} are not as substantial (a 1.7\% improvement over prior SOTA) using this approach. However, the 2-way model (from Table~\ref{2-way}) produced a $3.3\%$ improvement in \texttt{ANY-COMB}~F1-score. 

If we look at per-class F1-scores for non-\texttt{POS} classes, the 3-way model has an F1-score of $70.5\%$ for \texttt{NOCOMB}, but it only has an $F_1$ score of $26.3\%$ on \texttt{COMB} class; this is to be expected as \texttt{COMB} (non-positive combinations) is the least frequent class in the dataset (nearly 23\% of training data) and it is harder to distinguish it from the more frequent \texttt{POS} class.  Furthermore, Tiktinsky et al.~\cite{tiktinsky-etal-2022-dataset} also convey that for over 2/3 of the instances manual annotation necessitated consideration of additional context outside the sentence containing the drug names. But all results shown in Tables~\ref{2-way} and \ref{table:main-results} are trained on just the sentence containing the drug names. However, our results on including additional context from the enclosing full abstract context were underwhelming (more later). 

\section{Ablation and Error Analyses}
\label{sec-error}
\begin{table}[h]
\centering
 \renewcommand{\arraystretch}{1.2}
 \caption{\texttt{ANY-COMB} ablated $F_1$ scores on the test set.}
\label{tb:ablation}
\begin{tabular}{lccc}
\toprule
\textbf{Model} & \texttt{ANY-COMB} $F_1$ \\\midrule
\textbf{Full 2-way model}  & \textbf{72.7} \\
 \,\, w/o. 2-way classification   & 71.1 \\
 \,\, w/o. post-processing & 68.8 \\
  \,\, w/o. @DRUG@ entity type token & 69.5 \\
\bottomrule
\end{tabular}

\end{table}
To get at the relative importance of different ingredients of our models, we ran a small ablation experiment for the 2-way \texttt{ANY-COMB} model shown in Table~\ref{2-way} (2nd row). 
In Table~\ref{tb:ablation}, we observe that removing the 2-way approach and moving to a 3-way approach results in a $1.6\%$ dip in the score. If we remove the post-processing step involving hyphenated entities (as discussed in Section~\ref{sec-post}), the performance decreases by around $4\%$. This is not surprising because $10\%$ of entities contain at least one hyphen in the training data. The final row in Table~\ref{tb:ablation} shows the scores if we used a simple semicolon (;) symbol instead of the special @DRUG@ token to denote and separate drug entities in the linearization schema (Section~\ref{linearization}). Since there is only one entity type involved, we thought entity-specific tags may not be necessary. However, we see that using ``;'' instead of @DRUG@ lowers F1-score by over $3\%$. This is potentially due to the role semicolon plays in general English and overloading its functionality to also represent drug entities may have had unintended consequences in the final model.

We now discuss four different types of errors we often noticed in the model output.
\begin{enumerate}
    \item \textbf{Handling many-to-one attachments}: These are sentences where a drug is described to be used in combination with another drug from a list of candidates, leading to multiple combinations each with two drugs. In these cases, the language may be misinterpreted by our model to create larger combinations with more than two drugs. For example, consider the sentence: ``After successful phase II studies, recent phase III trials established combinations of \textbf{chlorambucil} with anti-CD20 antibodies such as \textbf{rituximab}, \textbf{ofatumumab} and \textbf{obinutuzumab} as a valuable treatment option for these patients.'' Here the gold prediction has three combinations each with two drugs: \textbf{chlorambucil} @DRUG@ \textbf{rituximab} @DRUG@  @POS@ \textbf{chlorambucil} @DRUG@ \textbf{ofatumumab} @DRUG@  @POS@ \textbf{chlorambucil} @DRUG@ \textbf{obinutuzumab} @DRUG@  @POS@. However, the model incorrecntly predicted a combination with all four drugs mentioned: \textbf{chlorambucil} @DRUG@ \textbf{rituximab} @DRUG@ \textbf{ofatumumab} @DRUG@ \textbf{obinutuzumab} @DRUG@ @POS@
    
    \item \textbf{Recognizing drugs that are not part of a combination}: There are occasions where the language contained in the sentence may not be conclusive enough to distinguish between a real combination in comparison with drugs that are being discussed as individual treatments. To drive this home, let's look at this sentence: ``\textbf{Dexamethasone} and \textbf{piroxicam} provided in the diet were found to significantly  inhibit lung tumors induced by 60 mg/kg vinyl carbamate at 24 weeks whereas \textbf{myo-inositol} also provided in the diet, did not significantly inhibit tumor  formation.''
    Here the gold annotation indicated that \textbf{Dexamethasone} and \textbf{piroxicam} are in a \texttt{POS} relation but our model predicted that all three drugs (including \textbf{myo-inositol}) are part of a size-3 \texttt{COMB} relation. The model was unable to notice that \textbf{myo-inositol} is not part of the combination. 

    \item \textbf{Distinguishing between \texttt{POS} and \texttt{COMB} relations}: At times, it was not straightforward from the intra-sentence context to determine whether a combination is effective or non-positive. Getting these relations right may need more complex multi-hop reasoning across different sentences of the abstract. Case in point is this sentence: ``Randomized trial of \textbf{lenalidomide} alone versus \textbf{lenalidomide} plus \textbf{rituximab}  in patients with recurrent follicular lymphoma : CALGB 50401 (Alliance).'' The gold annotation for this is a \texttt{POS} relation consisting of 
    \textbf{lenalidomide} and \textbf{rituximab}. However, the model predicted a \texttt{COMB} relation, which seems quite plausible when we examine the sentence. In fact, searching for this sentence on PubMed shows that this is the title of an article describing a clinical trial. The title does not give away enough information about whether the combination was effective or not, but the full abstract which was used by human annotators explicitly says that the combination is effective. 

     \item \textbf{Need for external domain knowledge}: Training data may  not have enough examples of indirect and implicit ways of communicating efficacy of medications. As such, external domain knowledge about biomedical concepts maybe needed to correctly capture certain relations. To demonstrate this, consider the sentence: ``{\color{blue} Growth inhibition} and {\color{blue}apoptosis} were significantly {\color{blue}higher} in BxPC-3, HPAC, and  PANC-1 cells treated with \textbf{celecoxib} and \textbf{erlotinib} than cells treated with either  \textbf{celecoxib} or \textbf{erlotinib}.'' Here efficacy is implicitly conveyed through the phrases ``growth inhibition'' and ``apoptosis'' and the quantifier ``higher''. The former two refer to concepts in disease mechanisms indicating that tumor cell production is lowered or tumor cells are dying, both implying a potential therapeutic effect. Thus, the gold \texttt{POS} relation between  \textbf{celecoxib} and \textbf{erlotinib} is missed by our model, which predicted as a \texttt{COMB} link. 
\end{enumerate}

\begin{table*}[!h]
\centering
\caption{Examples of the extended linearization schema for both NER and relation extraction  \label{tab:ner}}

\begin{tabular}{p{0.15\textwidth} p{0.58\textwidth}}
\toprule
Relations &
  Example  \\ \midrule
\multirow{3.5}{*}{\parbox{4cm}{Positive Combination}} &
  \textbf{Dexamethasone} and \textbf{piroxicam} provided in the diet were found to significantly  inhibit lung tumors induced by 60 mg/kg vinyl carbamate at 24 weeks whereas \textbf{myo-inositol} also provided in the diet, did not significantly inhibit tumor  formation. \\ \cmidrule(lr){2-2}
 &
 \texttt{{\textbf{Dexamethasone}; }{\textbf{piroxicam}; }{\textbf{myo-inositol} @NER@} {\textbf{Dexamethasone}; }{\textbf{piroxicam} @POS@}}  
   \\ \midrule
\multirow{3.5}{*}{\parbox{4cm}{Not a Combination}} &
  The results showed that \textbf{lamotrigine} did not produce any change in cognitive function, while \textbf{carbamazepine} produced cognitive dysfunction. \\ \cmidrule(lr){2-2}
 &
\texttt{{\textbf{lamotrigine}; }{\textbf{carbamazepine}} @NER@} 
   \\ \bottomrule
\end{tabular}
\vspace{-3mm}
\end{table*}

\section{Experiments with Longer Contexts}
\label{sec-long}
Thus far all our experiments were conducted using the input sentence that contained the drug names without considering any surrounding context from the enclosing full abstract. As was already alluded to in Section~\ref{sec-main-res}, 2/3 of the instances in \textbf{\texttt{CombDrugExt}} needed human annotators to consider other sentences outside the input sentence containing drug names~\cite{tiktinsky-etal-2022-dataset} to determine the correct labels. This behooves us to build models that consider $n$ sentences to the left and right of the main input sentence. 
We varied $n$ from 1 to 4 to better assess Seq2Rel's ability to work with a broader context. We added a special [SEP] token surrounding the relation-bearing sentence to inform the model that the task only considers drug entities and combinations expressed in the target sentence while it is still allowed to consider the neighboring context outside it. 

\begin{table}[h]
\centering
\renewcommand{\arraystretch}{1.3}
\caption{The effect of considering a context of $n$ sentences on either side of the relation-bearing sentence.}
\label{tb:long-context}
\begin{tabular}{lccc}
\toprule
Model & $F_1$ & $P$ & $R$\\\midrule
No additional context & \textbf{66.7} & 68.1 & 65.3 \\
 1 sentence of context & 58.3 & 63.2 & 54.0\\
 2 sentences of context & 62.8 & 68.5 & 58.0\\
 3 sentences of context & 65.7 & 69.9 & 62.0\\
 4 sentences of context & 66.2 & 67.8 & 64.7\\
\bottomrule
\end{tabular}

\end{table}

We present our evaluation results in Table~\ref{tb:long-context}, which clearly shows that considering additional context did not really help our situation. The first row (copy pasted from the last row of our main results  Table~\ref{table:main-results}) is clearly better than all other rows. A phenomenon that can be observed is that as $n$ goes from 1 to 4, the recall increases monotonously (and so does the F1-score). This gives us the impression that more context seems to help but it is still not able to beat the score when no additional context is used. This is in contrast with the findings of Tiktinsky et al.~\cite{tiktinsky-etal-2022-dataset} where extra context appeared to help. However, there is a crucial difference between their and our evaluation process --- the entities are provided and locked-in before the relations are predicted for them but our model is end-to-end and has to generate entities from the input. As such, the extra context may have confused our model in fetching the right entities, while it could have helped their span based approach with fixed entity locations.

\section{Drug NER Plus CST Experiment}

Please recall that the way the CDT extraction task was designed originally (Section~\ref{sec-formal}), the model only outputs drug entities that are part of a \texttt{POS} or \texttt{COMB} relations; drugs that are not part of any such relations are not output unless the input is a sentence where neither \texttt{POS} nor \texttt{COMB} relations occur. Thus named entity recognition (NER) of drugs in a comprehensive sense is not integral to the CDT task. This is a reasonable setting because in many cases, one may not be interested in drugs that do not form part of an interesting combination. However, there might be other scenarios where one may want to output all drug mentions even if they are not part of a relation. This could be for computing distributional statistics or to design more complex knowledge discovery applications. Since the annotation of \textbf{\texttt{CombDrugExt}} does contain all drug spans (even those that do not participate in a relation), we can design a extended CDT task where both entities and relations are required to be output (regardless of whether entities are part of a relation). This is the task we model in this section.

To also capture all drug entities, for \texttt{POS} and \texttt{COMB} relations, we extend the original linearization scheme by pre-pending all the entities mentioned in the sentence before relations are enumerated. For \texttt{NOCOMB} instances, only the entities are to be output (as the relation is self-explanatory when we do not see \texttt{POS} or \texttt{COMB} labels). Our extended   linearization scheme is shown in Table \ref{tab:ner}, where we can see the new special token @NER@ as a delimiter to indicate the end of the entity list before relation enumeration begins. We also replaced the @DRUG@ token with a semicolon here for simplification, especially since we are already listing all drugs upfront. The first \texttt{POS} example shown in the table has three drugs listed for NER using this extended schema;  \textbf{myo-inositol} would not have been part of the output in the original schema.

From Table~\ref{tb:nerf1}, we see that a Seq2Rel model trained with this updated linearization schema (for both NER and CDT) achieves an F1-score of $94\%$ for drug NER.  
However, there is a dip of around $1\%$ in \texttt{POS} F1-score compared to the approach without the NER output (last row of Table~\ref{table:main-results}). But surprisingly the \texttt{ANY-COMB} F1-score is clearly better (by at least $1.3\%$) than the corresponding scores obtained through both 2-way and 3-way models as per the original approach (from Table~\ref{2-way}). Given the ability to achieve high NER score, along with high scores for \texttt{POS} (within $1\%$ of SoTA) and \texttt{ANY-COMB} (new SoTA), we believe this approach of jointly accounting for both entities and relations is better than the original approach that outputs only relations.

\begin{table}[h]
\centering
 \renewcommand{\arraystretch}{1.3}
 \caption{End-to-end drug NER and CDT extraction precision, recall, and $F_1$-scores for \texttt{POS} and \texttt{ANY-COMB} using the linearization schema from Table~\ref{tab:ner}. }
\label{tb:nerf1}
\begin{tabular}{lccc}
\toprule
Task & $F_1$ & P & R\\\midrule
NER  & 94.0 & 93.5 & 94.5 \\\midrule
 \texttt{POS} Combination CDT & 65.8 & 64.9 & 66.7 \\
 Any Combination CDT & 74.0 & 75.9 & 72.2\\
\bottomrule
\end{tabular}

\end{table}

\section{Left-to-right linearization schema}
\label{sec-left-to-right}
In Section~\ref{linearization}, linearization simply followed the order in which the drug spans and combinations were ordered by the creators of \textbf{\texttt{CombDrugExt}}. While this order was often the left-to-right order in the input sentence, this was not always the case.  However, the original Seq2Rel method~\cite{giorgi-etal-2022-sequence} uses a strict left-to-right ordering of entity spans as observed in the input sentence during training time. Of course, at test time, ordering is irrelevant as we are matching sets (not lists). We also experimented with this left-to-right ordering.  
That is, in the linearized output, a combination involving drugs that appear  earlier in the sentence appears before another combination whose constituent drugs appear later. If two combinations share a drug, the order of the remaining unshared drugs in the input sentence determines combination order in the output\footnote{More formally, the drug spans form a finite total order based on their left to right positioning in the input sentence. Our linearization strategy is based on the well-known extension of a total order on a finite set (drug spans) to a total order on its power set (drug combinations).}. The results were very similar to results showed thus far in the paper with minor variations: $<1\%$  performance shifts that are not always in favor of the original or left-to-right order. So we believe these variations are not worth discussing in the rest of this paper.  

\section{Related Work}
Biomedical relation extraction (RE)  has mostly focused on a sub-problem, where the entities are assumed to be already provided, making it a relation classification problem~\cite{rink2011automatic,kavuluru2017extracting,liu2016dependency,peng2018extracting}. In the end-to-end setting, raw textual input is provided and the task is to simultaneously output both entities and relations. While there is merit in considering non end-to-end settings in complex RE problems, overall the end-to-end setup is more realistic for assessment and evaluation for application design.
There has been a recent surge in end-to-end approaches in the general RE domain. These can be classified into two main types: (1) A pipeline approach that feeds the output of an NER model to a relation classification model~\cite{zhong-chen-2021-frustratingly}. (2) A joint modeling approach that simultaneously extracts both entities and relations~\cite{miwa2016end,tran2019neural,bekoulis2018joint}.  
In particular, sequence-to-sequence models for end-to-end relation extraction have been on the rise~\cite{giorgi-etal-2022-sequence,zeng-etal-2018-extracting,zeng2020copymtl,nayak2020effective} that tend to fall in the joint modeling group of methods. Especially, these approaches easily lend themselves to the dynamic $n$-ary nature of the CDT extraction task and hence we employ them in this paper. 

There are few efforts in biomedical natural language processing that handle drug-disease treatment relations. Kilicoglu et al.~\cite{kilicoglu2020broad} present a broad coverage hybrid system to extract different types of relations, which also includes treatment relations. Dumitrache et al.~\cite{dumitrache2018crowdsourcing} release one of the few publicly available datasets with both expert annotated and crowd sourced treatment relation annotations. $n$-ary relation extraction datasets are also not widely available in biomedicine with the major exception of drug-gene-mutation relations~\cite{peng-etal-2017-cross}.  Finally, Tiktinsky et al.~\cite{tiktinsky-etal-2022-dataset} are the first to annotate and publicly release a dataset for drug combinations, which forms the main focus of our manuscript.

\section{Conclusion}
In this study, we adapted an end-to-end relation extraction approach  based on sequence-to-sequence methods for extracting drug combinations that are part of combination drug therapies. The resulting Seq2Rel models use linearization schemes to encode entities and relations as output sequences for training the models and to use them in the generation mode for testing. We showed that this approach results in new state-of-the-art performances on the \textbf{\texttt{CombDrugExt}} dataset in the end-to-end setting that improve even over non end-to-end baselines (by nearly $5\%$ in F1-score), established by prior efforts. We see five important directions for future efforts:
\begin{itemize}
\item As highlighted in Section~\ref{sec-classes}, the \textbf{\texttt{CombDrugExt}} dataset does not include the target disease, which we believe is central to any future work on this problem. We propose to extend textbf{\texttt{CombDrugExt}}  by specifically adding this extra piece of information to make it a more complete dataset for CDT extraction. 

\item As per our results in Section~\ref{sec-long}, our method, as it stands, is unable to take advantage of the additional full abstract context available. We plan  to further adapt Seq2Rel methods to take advantage of potential multi-hop links that are present in the broader context of the enclosing abstract. Based on manual examination of some errors, we believe co-references of the participating drug combinations may need to be handled more carefully. 

\item A minor issue with order selection in linearization (Sections~\ref{linearization} and \ref{sec-left-to-right}) is that during training time, we unfairly penalize the target tokens if they do not exactly match the order we specify, even if they are in essence representing the same combination. We believe that an ensemble model where each constituent model trains with a   different valid output order that nevertheless encodes the same combinations may perform much better than any single model trained with a fixed order. This is a hypothesis we aim to test in the future. 

\item It may be important to qualify our end-to-end formulation in Section~\ref{sec-formal} with a caveat. Although we do not assume that the drug spans are available upfront, we still assume that the sentence that contains the names of all participating drugs is available. Ideally, for CDT  extraction to be truly end-to-end, our method ought to extract relations given the full abstract without any index of the sentence containing the potential combination. A future direction would be to operate without the knowledge of the sentence containing the drug names. However, this might need additional human curated annotations and may be a more expensive process.

\item In Section~\ref{sec-error} we identified different types of errors, one of which was due to the dependence on domain knowledge about treatment mechanisms through which drugs operate; we surmised that this knowledge may not have been available from the language variation in the training examples.  A future direction would be to imbue this domain knowledge into the modeling process for CDT extraction, potentially through external knowledge graphs of biomedical processes. Of course, in this case, for fair comparison, this external knowledge should also be provided in some manner along with the dataset for benchmarking by the wider community. 
\end{itemize}
\newpage
\bibliographystyle{IEEEtran}
\bibliography{IEEEabrv,bibtex}
\end{document}